\documentclass[11pt,a4paper]{article}
\usepackage[hyperref]{acl2019}

\usepackage{url}
\usepackage{array}
\usepackage{times}
\usepackage{helvet}
\usepackage{amssymb}
\usepackage{courier}
\usepackage{graphicx}

\aclfinalcopy
\def\aclpaperid{423}
\title{Explicit Utilization of General Knowledge \\ in Machine Reading Comprehension}

\author{
Chao Wang and Hui Jiang \\
Department of Electrical Engineering and Computer Science \\
Lassonde School of Engineering, York University \\
4700 Keele Street, Toronto, Ontario, Canada \\
{\tt $\{chwang, hj\}$@eecs.yorku.ca}
}

\date{}

\begin{document}
\maketitle
\begin{abstract}
To bridge the gap between Machine Reading Comprehension (MRC) models and human beings, which is mainly reflected in the hunger for data and the robustness to noise, in this paper, we explore how to integrate the neural networks of MRC models with the general knowledge of human beings. On the one hand, we propose a data enrichment method, which uses WordNet to extract inter-word semantic connections as general knowledge from each given passage-question pair. On the other hand, we propose an end-to-end MRC model named as Knowledge Aided Reader (KAR), which explicitly uses the above extracted general knowledge to assist its attention mechanisms. Based on the data enrichment method, KAR is comparable in performance with the state-of-the-art MRC models, and significantly more robust to noise than them. When only a subset ($20\%$--$80\%$) of the training examples are available, KAR outperforms the state-of-the-art MRC models by a large margin, and is still reasonably robust to noise.
\end{abstract}

\section{Introduction}
\begin{table*}
\centering
\begin{tabular}
{|m{0.5\linewidth}|m{0.3\linewidth}|m{0.11\linewidth}|}
\hline
\textbf{Passage} &
\textbf{Question} &
\textbf{Answer} \\
\hline
Teachers may use a lesson plan to \textbf{facilitate} student learning, providing a course of study which is called the curriculum. &
What can a teacher use to \textbf{help} students learn? &
lesson plan \\
\hline
Manufacturing accounts for a significant but declining share of employment, although the city's garment industry is showing a resurgence in \textbf{Brooklyn}. &
In what \textbf{borough} is the garment business prominent? &
Brooklyn \\
\hline
\end{tabular}
\caption{\label{t1} Two examples about the importance of inter-word semantic connections to the reading comprehension ability of human beings: in the first one, we can find the answer because we know ``facilitate'' is a synonym of ``help''; in the second one, we can find the answer because we know ``Brooklyn'' is a hyponym of ``borough''.}
\end{table*}
Machine Reading Comprehension (MRC), as the name suggests, requires a machine to read a passage and answer its relevant questions. Since the answer to each question is supposed to stem from the corresponding passage, a common MRC solution is to develop a neural-network-based MRC model that predicts an answer span (i.e. the answer start position and the answer end position) from the passage of each given passage-question pair. To facilitate the explorations and innovations in this area, many MRC datasets have been established, such as SQuAD \cite{rajpurkarpranav:2016}, MS MARCO \cite{nguyentri:2016}, and TriviaQA \cite{joshimandar:2017}. Consequently, many pioneering MRC models have been proposed, such as BiDAF \cite{seominjoon:2016}, R-NET \cite{wangwenhui:2017}, and QANet \cite{yuadamswei:2018}. According to the leader board of SQuAD, the state-of-the-art MRC models have achieved the same performance as human beings. However, does this imply that they have possessed the same reading comprehension ability as human beings? \\
OF COURSE NOT. There is a huge gap between MRC models and human beings, which is mainly reflected in the hunger for data and the robustness to noise. On the one hand, developing MRC models requires a large amount of training examples (i.e. the passage-question pairs labeled with answer spans), while human beings can achieve good performance on evaluation examples (i.e. the passage-question pairs to address) without training examples. On the other hand, \citet{jiarobin:2017} revealed that intentionally injected noise (e.g. misleading sentences) in evaluation examples causes the performance of MRC models to drop significantly, while human beings are far less likely to suffer from this. The reason for these phenomena, we believe, is that MRC models can only utilize the knowledge contained in each given passage-question pair, but in addition to this, human beings can also utilize general knowledge. A typical category of general knowledge is inter-word semantic connections. As shown in Table~\ref{t1}, such general knowledge is essential to the reading comprehension ability of human beings. \\
A promising strategy to bridge the gap mentioned above is to integrate the neural networks of MRC models with the general knowledge of human beings. To this end, it is necessary to solve two problems: extracting general knowledge from passage-question pairs and utilizing the extracted general knowledge in the prediction of answer spans. The first problem can be solved with knowledge bases, which store general knowledge in structured forms. A broad variety of knowledge bases are available, such as WordNet \cite{fellbaumchristiane:1998} storing semantic knowledge, ConceptNet \cite{speerrobert:2017} storing commonsense knowledge, and Freebase \cite{bollackerkurt:2008} storing factoid knowledge. In this paper, we limit the scope of general knowledge to inter-word semantic connections, and thus use WordNet as our knowledge base. The existing way to solve the second problem is to encode general knowledge in vector space so that the encoding results can be used to enhance the lexical or contextual representations of words \cite{weissenborndirk:2017, mihaylovtodor:2018}. However, this is an implicit way to utilize general knowledge, since in this way we can neither understand nor control the functioning of general knowledge. In this paper, we discard the existing implicit way and instead explore an explicit (i.e. understandable and controllable) way to utilize general knowledge. \\
The contribution of this paper is two-fold. On the one hand, we propose a data enrichment method, which uses WordNet to extract inter-word semantic connections as general knowledge from each given passage-question pair. On the other hand, we propose an end-to-end MRC model named as Knowledge Aided Reader (KAR), which explicitly uses the above extracted general knowledge to assist its attention mechanisms. Based on the data enrichment method, KAR is comparable in performance with the state-of-the-art MRC models, and significantly more robust to noise than them. When only a subset ($20\%$--$80\%$) of the training examples are available, KAR outperforms the state-of-the-art MRC models by a large margin, and is still reasonably robust to noise.

\section{Data Enrichment Method}
In this section, we elaborate a WordNet-based data enrichment method, which is aimed at extracting inter-word semantic connections from each passage-question pair in our MRC dataset. The extraction is performed in a controllable manner, and the extracted results are provided as general knowledge to our MRC model.

\subsection{Semantic Relation Chain}
WordNet is a lexical database of English, where words are organized into synsets according to their senses. A synset is a set of words expressing the same sense so that a word having multiple senses belongs to multiple synsets, with each synset corresponding to a sense. Synsets are further related to each other through semantic relations. According to the WordNet interface provided by NLTK \cite{birdsteven:2004}, there are totally sixteen types of semantic relations (e.g. hypernyms, hyponyms, holonyms, meronyms, attributes, etc.). Based on synset and semantic relation, we define a new concept: semantic relation chain. A semantic relation chain is a concatenated sequence of semantic relations, which links a synset to another synset. For example, the synset ``keratin.n.01'' is related to the synset ``feather.n.01'' through the semantic relation ``substance holonym'', the synset ``feather.n.01'' is related to the synset ``bird.n.01'' through the semantic relation ``part holonym'', and the synset ``bird.n.01'' is related to the synset ``parrot.n.01'' through the semantic relation ``hyponym'', thus ``substance holonym $\to$ part holonym $\to$ hyponym'' is a semantic relation chain, which links the synset ``keratin.n.01'' to the synset ``parrot.n.01''. We name each semantic relation in a semantic relation chain as a hop, therefore the above semantic relation chain is a $3$-hop chain. By the way, each single semantic relation is equivalent to a $1$-hop chain.

\subsection{Inter-word Semantic Connection}
The key problem in the data enrichment method is determining whether a word is semantically connected to another word. If so, we say that there exists an inter-word semantic connection between them. To solve this problem, we define another new concept: the extended synsets of a word. Given a word $w$, whose synsets are represented as a set $S_w$, we use another set $S^*_w$ to represent its extended synsets, which includes all the synsets that are in $S_w$ or that can be linked to from $S_w$ through semantic relation chains. Theoretically, if there is no limitation on semantic relation chains, $S^*_w$ will include all the synsets in WordNet, which is meaningless in most situations. Therefore, we use a hyper-parameter $\kappa \in \mathbb{N}$ to represent the permitted maximum hop count of semantic relation chains. That is to say, only the chains having no more than $\kappa$ hops can be used to construct $S^*_w$ so that $S^*_w$ becomes a function of $\kappa$: $S^*_w(\kappa)$ (if $\kappa = 0$, we will have $S^*_w(0) = S_w$). Based on the above statements, we formulate a heuristic rule for determining inter-word semantic connections: a word $w_1$ is semantically connected to another word $w_2$ if and only if $S^*_{w_1}(\kappa) \cap S_{w_2} \ne \emptyset$.

\subsection{General Knowledge Extraction}
Given a passage-question pair, the inter-word semantic connections that connect any word to any passage word are regarded as the general knowledge we need to extract. Considering the requirements of our MRC model, we only extract the positional information of such inter-word semantic connections. Specifically, for each word $w$, we extract a set $E_w$, which includes the positions of the passage words that $w$ is semantically connected to (if $w$ itself is a passage word, we will exclude its own position from $E_w$). We can control the amount of the extracted results by setting the hyper-parameter $\kappa$: if we set $\kappa$ to $0$, inter-word semantic connections will only exist between synonyms; if we increase $\kappa$, inter-word semantic connections will exist between more words. That is to say, by increasing $\kappa$ within a certain range, we can usually extract more inter-word semantic connections from a passage-question pair, and thus can provide the MRC model with more general knowledge. However, due to the complexity and diversity of natural languages, only a part of the extracted results can serve as useful general knowledge, while the rest of them are useless for the prediction of answer spans, and the proportion of the useless part always rises when $\kappa$ is set larger. Therefore we set $\kappa$ through cross validation (i.e. according to the performance of the MRC model on the development examples).

\section{Knowledge Aided Reader}
In this section, we elaborate our MRC model: Knowledge Aided Reader (KAR). The key components of most existing MRC models are their attention mechanisms \cite{bahdanaudzmitry:2014}, which are aimed at fusing the associated representations of each given passage-question pair. These attention mechanisms generally fall into two categories: the first one, which we name as mutual attention, is aimed at fusing the question representations into the passage representations so as to obtain the question-aware passage representations; the second one, which we name as self attention, is aimed at fusing the question-aware passage representations into themselves so as to obtain the final passage representations. Although KAR is equipped with both categories, its most remarkable feature is that it explicitly uses the general knowledge extracted by the data enrichment method to assist its attention mechanisms. Therefore we separately name the attention mechanisms of KAR as knowledge aided mutual attention and knowledge aided self attention.

\subsection{Task Definition}
Given a passage $P = \{p_1, \ldots, p_n\}$ and a relevant question $Q = \{q_1, \ldots, q_m\}$, the task is to predict an answer span $[a_s, a_e]$, where $1 \le a_s \le a_e \le n$, so that the resulting subsequence $\{p_{a_s}, \ldots, p_{a_e}\}$ from $P$ is an answer to $Q$.

\subsection{Overall Architecture}
\begin{figure*}
\centering
\includegraphics[width=0.98\linewidth]{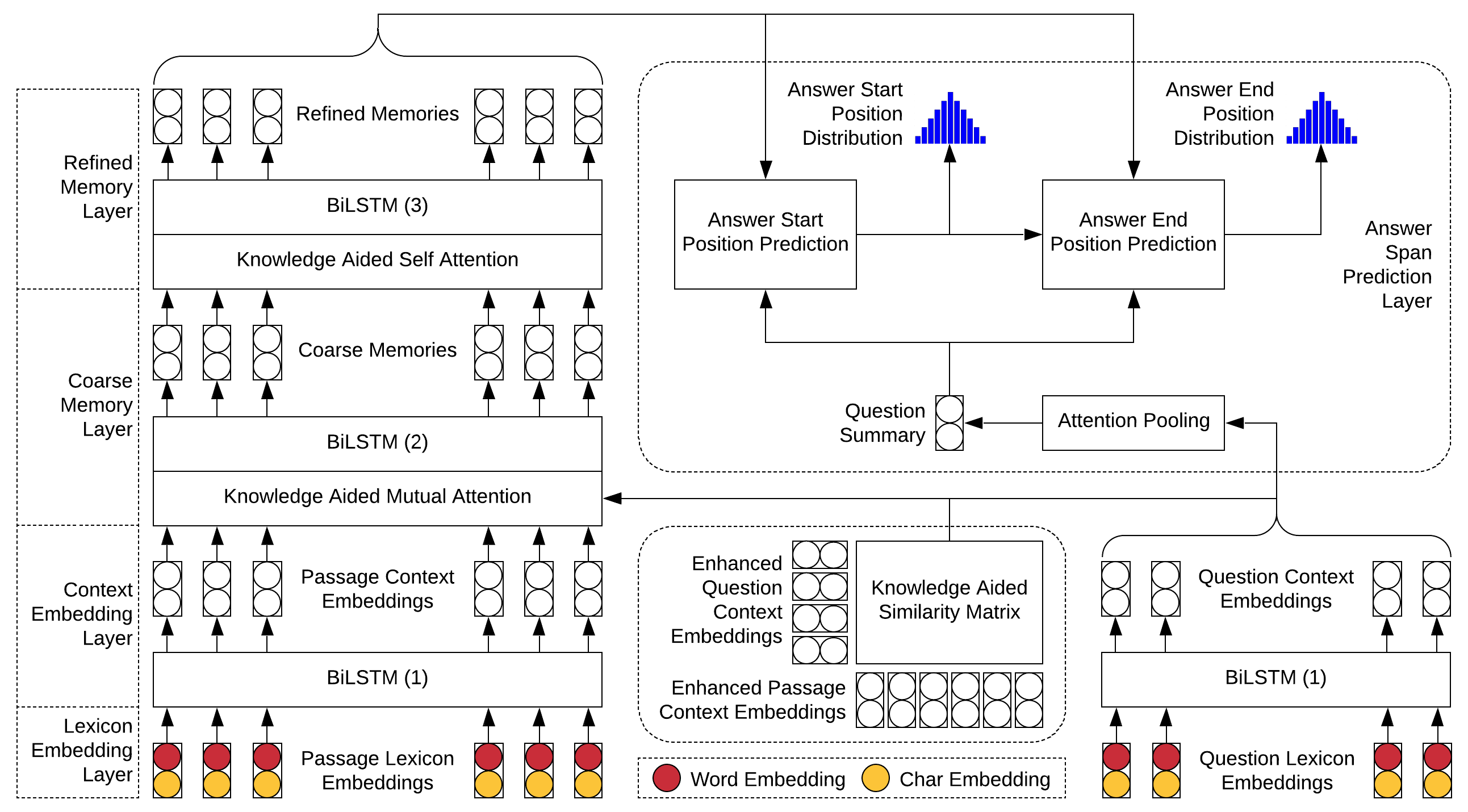}
\caption{\label{f1} An end-to-end MRC model: Knowledge Aided Reader (KAR)}
\end{figure*}
As shown in Figure~\ref{f1}, KAR is an end-to-end MRC model consisting of five layers: \\
\textbf{Lexicon Embedding Layer.} This layer maps the words to the lexicon embeddings. The lexicon embedding of each word is composed of its word embedding and character embedding. For each word, we use the pre-trained GloVe \cite{penningtonjeffrey:2014} word vector as its word embedding, and obtain its character embedding with a Convolutional Neural Network (CNN) \cite{kimyoon:2014}. For both the passage and the question, we pass the concatenation of the word embeddings and the character embeddings through a shared dense layer with ReLU activation, whose output dimensionality is $d$. Therefore we obtain the passage lexicon embeddings $L_P \in \mathbb{R}^{d \times n}$ and the question lexicon embeddings $L_Q \in \mathbb{R}^{d \times m}$. \\
\textbf{Context Embedding Layer.} This layer maps the lexicon embeddings to the context embeddings. For both the passage and the question, we process the lexicon embeddings (i.e. $L_P$ for the passage and $L_Q$ for the question) with a shared bidirectional LSTM (BiLSTM) \cite{hochreitersepp:1997}, whose hidden state dimensionality is $\frac{1}{2}d$. By concatenating the forward LSTM outputs and the backward LSTM outputs, we obtain the passage context embeddings $C_P \in \mathbb{R}^{d \times n}$ and the question context embeddings $C_Q \in \mathbb{R}^{d \times m}$. \\
\textbf{Coarse Memory Layer.} This layer maps the context embeddings to the coarse memories. First we use knowledge aided mutual attention (introduced later) to fuse $C_Q$ into $C_P$, the outputs of which are represented as $\tilde{G} \in \mathbb{R}^{d \times n}$. Then we process $\tilde{G}$ with a BiLSTM, whose hidden state dimensionality is $\frac{1}{2}d$. By concatenating the forward LSTM outputs and the backward LSTM outputs, we obtain the coarse memories $G \in \mathbb{R}^{d \times n}$, which are the question-aware passage representations. \\
\textbf{Refined Memory Layer.} This layer maps the coarse memories to the refined memories. First we use knowledge aided self attention (introduced later) to fuse $G$ into themselves, the outputs of which are represented as $\tilde{H} \in \mathbb{R}^{d \times n}$. Then we process $\tilde{H}$ with a BiLSTM, whose hidden state dimensionality is $\frac{1}{2}d$. By concatenating the forward LSTM outputs and the backward LSTM outputs, we obtain the refined memories $H \in \mathbb{R}^{d \times n}$, which are the final passage representations. \\
\textbf{Answer Span Prediction Layer.} This layer predicts the answer start position and the answer end position based on the above layers. First we obtain the answer start position distribution $o_s$:
\[t_i = v_s^\top \mathrm{tanh}(W_s h_{p_i} + U_s r_Q) \in \mathbb{R}\]
\[o_s = \mathrm{softmax}(\{t_1, \ldots, t_n\}) \in \mathbb{R}^n\]
where $v_s$, $W_s$, and $U_s$ are trainable parameters; $h_{p_i}$ represents the refined memory of each passage word $p_i$ (i.e. the $i$-th column in $H$); $r_Q$ represents the question summary obtained by performing an attention pooling over $C_Q$. Then we obtain the answer end position distribution $o_e$:
\[t_i = v_e^\top \mathrm{tanh}(W_e h_{p_i} + U_e [r_Q; H o_s]) \in \mathbb{R}\]
\[o_e = \mathrm{softmax}(\{t_1, \ldots, t_n\}) \in \mathbb{R}^n\]
where $v_e$, $W_e$, and $U_e$ are trainable parameters; $[;]$ represents vector concatenation. Finally we construct an answer span prediction matrix $O = \mathrm{uptri}(o_s o_e^\top) \in \mathbb{R}^{n \times n}$, where $\mathrm{uptri}(X)$ represents the upper triangular matrix of a matrix $X$. Therefore, for the training, we minimize $-\mathrm{log}(O_{a_s, a_e})$ on each training example whose labeled answer span is $[a_s, a_e]$; for the inference, we separately take the row index and column index of the maximum element in $O$ as $a_s$ and $a_e$.

\subsection{Knowledge Aided Mutual Attention}
As a part of the coarse memory layer, knowledge aided mutual attention is aimed at fusing the question context embeddings $C_Q$ into the passage context embeddings $C_P$, where the key problem is to calculate the similarity between each passage context embedding $c_{p_i}$ (i.e. the $i$-th column in $C_P$) and each question context embedding $c_{q_j}$ (i.e. the $j$-th column in $C_Q$). To solve this problem, \citet{seominjoon:2016} proposed a similarity function:
\[f(c_{p_i}, c_{q_j}) = v_f^\top [c_{p_i}; c_{q_j}; c_{p_i} \odot c_{q_j}] \in \mathbb{R}\]
where $v_f$ is a trainable parameter; $\odot$ represents element-wise multiplication. This similarity function has also been adopted by several other works \cite{clarkchristopher:2017,yuadamswei:2018}. However, since context embeddings contain high-level information, we believe that introducing the pre-extracted general knowledge into the calculation of such similarities will make the results more reasonable. Therefore we modify the above similarity function to the following form:
\[f^*(c_{p_i}, c_{q_j}) = v_f^\top [c^*_{p_i}; c^*_{q_j}; c^*_{p_i} \odot c^*_{q_j}] \in \mathbb{R}\]
where $c^*_x$ represents the enhanced context embedding of a word $x$. We use the pre-extracted general knowledge to construct the enhanced context embeddings. Specifically, for each word $w$, whose context embedding is $c_w$, to construct its enhanced context embedding $c^*_w$, first recall that we have extracted a set $E_w$, which includes the positions of the passage words that $w$ is semantically connected to, thus by gathering the columns in $C_P$ whose indexes are given by $E_w$, we obtain the matching context embeddings $Z \in \mathbb{R}^{d \times |E_w|}$. Then by constructing a $c_w$-attended summary of $Z$, we obtain the matching vector $c^+_w$ (if $E_w = \emptyset$, which makes $Z = \{\}$, we will set $c^+_w = 0$):
\[t_i = v_c^\top \mathrm{tanh}(W_c z_i + U_c c_w) \in \mathbb{R}\]
\[c^+_w = Z \ \mathrm{softmax}(\{t_1, \ldots, t_{|E_w|}\}) \in \mathbb{R}^{d}\]
where $v_c$, $W_c$, and $U_c$ are trainable parameters; $z_i$ represents the $i$-th column in $Z$. Finally we pass the concatenation of $c_w$ and $c^+_w$ through a dense layer with ReLU activation, whose output dimensionality is $d$. Therefore we obtain the enhanced context embedding $c^*_w \in \mathbb{R}^{d}$. \\
Based on the modified similarity function and the enhanced context embeddings, to perform knowledge aided mutual attention, first we construct a knowledge aided similarity matrix $A \in \mathbb{R}^{n \times m}$, where each element $A_{i,j} = f^*(c_{p_i}, c_{q_j})$. Then following \citet{yuadamswei:2018}, we construct the passage-attended question summaries $R_Q$ and the question-attended passage summaries $R_P$:
\[R_Q = C_Q \ \mathrm{softmax}_r^\top(A) \in \mathbb{R}^{d \times n}\]
\[R_P = C_P \ \mathrm{softmax}_c(A) \ \mathrm{softmax}_r^\top(A) \in \mathbb{R}^{d \times n}\]
where $\mathrm{softmax}_r$ represents softmax along the row dimension and $\mathrm{softmax}_c$ along the column dimension. Finally following \citet{clarkchristopher:2017}, we pass the concatenation of $C_P$, $R_Q$, $C_P \odot R_Q$, and $R_P \odot R_Q$ through a dense layer with ReLU activation, whose output dimensionality is $d$. Therefore we obtain the outputs $\tilde{G} \in \mathbb{R}^{d \times n}$.

\subsection{Knowledge Aided Self Attention}
As a part of the refined memory layer, knowledge aided self attention is aimed at fusing the coarse memories $G$ into themselves. If we simply follow the self attentions of other works \cite{wangwenhui:2017,huanghsinyuan:2017,liuxiaodong:2017,clarkchristopher:2017}, then for each passage word $p_i$, we should fuse its coarse memory $g_{p_i}$ (i.e. the $i$-th column in $G$) with the coarse memories of all the other passage words. However, we believe that this is both unnecessary and distracting, since each passage word has nothing to do with many of the other passage words. Thus we use the pre-extracted general knowledge to guarantee that the fusion of coarse memories for each passage word will only involve a precise subset of the other passage words. Specifically, for each passage word $p_i$, whose coarse memory is $g_{p_i}$, to perform the fusion of coarse memories, first recall that we have extracted a set $E_{p_i}$, which includes the positions of the other passage words that $p_i$ is semantically connected to, thus by gathering the columns in $G$ whose indexes are given by $E_{p_i}$, we obtain the matching coarse memories $Z \in \mathbb{R}^{d \times |E_{p_i}|}$. Then by constructing a $g_{p_i}$-attended summary of $Z$, we obtain the matching vector $g^+_{p_i}$ (if $E_{p_i} = \emptyset$, which makes $Z = \{\}$, we will set $g^+_{p_i} = 0$):
\[t_i = v_g^\top \mathrm{tanh}(W_g z_i + U_g g_{p_i}) \in \mathbb{R}\]
\[g^+_{p_i} = Z \ \mathrm{softmax}(\{t_1, \ldots, t_{|E_{p_i}|}\}) \in \mathbb{R}^{d}\]
where $v_g$, $W_g$, and $U_g$ are trainable parameters. Finally we pass the concatenation of $g_{p_i}$ and $g^+_{p_i}$ through a dense layer with ReLU activation, whose output dimensionality is $d$. Therefore we obtain the fusion result $\tilde{h}_{p_i} \in \mathbb{R}^{d}$, and further the outputs $\tilde{H} = \{\tilde{h}_{p_1}, \ldots, \tilde{h}_{p_n}\} \in \mathbb{R}^{d \times n}$.

\section{Related Works}
\textbf{Attention Mechanisms.} Besides those mentioned above, other interesting attention mechanisms include performing multi-round alignment to avoid the problems of attention redundancy and attention deficiency \cite{huminghao:2017}, and using mutual attention as a skip-connector to densely connect pairwise layers \cite{tayyi:2018}. \\
\textbf{Data Augmentation.} It is proved that properly augmenting training examples can improve the performance of MRC models. For example, \citet{yangzhilin:2017} trained a generative model to generate questions based on unlabeled text, which substantially boosted their performance; \citet{yuadamswei:2018} trained a back-and-forth translation model to paraphrase training examples, which brought them a significant performance gain. \\
\textbf{Multi-step Reasoning.} Inspired by the fact that human beings are capable of understanding complex documents by reading them over and over again, multi-step reasoning was proposed to better deal with difficult MRC tasks. For example, \citet{shenyelong:2017} used reinforcement learning to dynamically determine the number of reasoning steps; \citet{liuxiaodong:2017} fixed the number of reasoning steps, but used stochastic dropout in the output layer to avoid step bias. \\
\textbf{Linguistic Embeddings.} It is both easy and effective to incorporate linguistic embeddings into the input layer of MRC models. For example, \citet{chendanqi:2017} and \citet{liuxiaodong:2017} used POS embeddings and NER embeddings to construct their input embeddings; \citet{liurui:2017} used structural embeddings based on parsing trees to constructed their input embeddings. \\
\textbf{Transfer Learning.} Several recent breakthroughs in MRC benefit from feature-based transfer learning \cite{mccannbryan:2017,petersmatthewe:2018} and fine-tuning-based transfer learning \cite{radfordalec:2018,devlinjacob:2018}, which are based on certain word-level or sentence-level models pre-trained on large external corpora in certain supervised or unsupervised manners.

\section{Experiments}
\subsection{Experimental Settings}
\textbf{MRC Dataset.} The MRC dataset used in this paper is SQuAD 1.1, which contains over $100,000$ passage-question pairs and has been randomly partitioned into three parts: a training set ($80\%$), a development set ($10\%$), and a test set ($10\%$). Besides, we also use two of its adversarial sets, namely AddSent and AddOneSent \cite{jiarobin:2017}, to evaluate the robustness to noise of MRC models. The passages in the adversarial sets contain misleading sentences, which are aimed at distracting MRC models. Specifically, each passage in AddSent contains several sentences that are similar to the question but not contradictory to the answer, while each passage in AddOneSent contains a human-approved random sentence that may be unrelated to the passage. \\
\textbf{Implementation Details.} We tokenize the MRC dataset with spaCy 2.0.13 \cite{honnibalmatthew:2017}, manipulate WordNet 3.0 with NLTK 3.3, and implement KAR with TensorFlow 1.11.0 \cite{abadimartin:2016}. For the data enrichment method, we set the hyper-parameter $\kappa$ to $3$. For the dense layers and the BiLSTMs, we set the dimensionality unit $d$ to $600$. For model optimization, we apply the Adam \cite{kingmadiederikp:2014} optimizer with a learning rate of $0.0005$ and a mini-batch size of $32$. For model evaluation, we use Exact Match (EM) and F1 score as evaluation metrics. To avoid overfitting, we apply dropout \cite{srivastavanitish:2014} to the dense layers and the BiLSTMs with a dropout rate of $0.3$. To boost the performance, we apply exponential moving average with a decay rate of $0.999$.

\subsection{Model Comparison in both Performance and the Robustness to Noise}
\begin{table*}
\centering
\begin{tabular}
{|m{0.34\linewidth}|m{0.13\linewidth}|m{0.13\linewidth}|m{0.13\linewidth}|m{0.13\linewidth}|}
\hline
\textbf{Single MRC model} &
\textbf{Dev set \newline (EM / F1)} &
\textbf{Test set \newline (EM / F1)} &
\textbf{AddSent \newline (F1)} &
\textbf{AddOneSent \newline (F1)} \\
\hline
FusionNet \cite{huanghsinyuan:2017} & 75.3 / 83.6 & 76.0 / 83.9 & 51.4 & 60.7 \\
\hline
RaSoR+TR+LM \cite{salantshimi:2017} & 77.0 / 84.0 & 77.6 / 84.2 & 47.0 & 57.0 \\
\hline
SAN \cite{liuxiaodong:2017} & 76.2 / 84.1 & 76.8 / 84.4 & 46.6 & 56.5 \\
\hline
R.M-Reader \cite{huminghao:2017} & \textbf{78.9 / 86.3} & 79.5 / 86.6 & 58.5 & 67.0 \\
\hline
QANet (with data augmentation) \cite{yuadamswei:2018} & 75.1 / 83.8 & \textbf{82.5 / 89.3} & 45.2 & 55.7 \\
\hline
\textbf{KAR (ours)} & 76.7 / 84.9 & 76.1 / 83.5 & \textbf{60.1} & \textbf{72.3} \\
\hline
\end{tabular}
\caption{\label{t2} Model comparison based on SQuAD 1.1 and two of its adversarial sets: AddSent and AddOneSent. All the numbers are up to date as of October 18, 2018. Note that SQuAD 2.0 \cite{rajpurkarpranav:2018} is not involved in this paper, because it requires MRC models to deal with the problem of answer triggering, but this paper is aimed at improving the hunger for data and robustness to noise of MRC models.}
\end{table*}
We compare KAR with other MRC models in both performance and the robustness to noise. Specifically, we not only evaluate the performance of KAR on the development set and the test set, but also do this on the adversarial sets. As for the comparative objects, we only consider the single MRC models that rank in the top 20 on the SQuAD 1.1 leader board and have reported their performance on the adversarial sets. There are totally five such comparative objects, which can be considered as representatives of the state-of-the-art MRC models. As shown in Table~\ref{t2}, on the development set and the test set, the performance of KAR is on par with that of the state-of-the-art MRC models; on the adversarial sets, KAR outperforms the state-of-the-art MRC models by a large margin. That is to say, KAR is comparable in performance with the state-of-the-art MRC models, and significantly more robust to noise than them. \\
To verify the effectiveness of general knowledge, we first study the relationship between the amount of general knowledge and the performance of KAR. As shown in Table~\ref{t3}, by increasing $\kappa$ from $0$ to $5$ in the data enrichment method, the amount of general knowledge rises monotonically, but the performance of KAR first rises until $\kappa$ reaches $3$ and then drops down. Then we conduct an ablation study by replacing the knowledge aided attention mechanisms with the mutual attention proposed by \citet{seominjoon:2016} and the self attention proposed by \citet{wangwenhui:2017} separately, and find that the F1 score of KAR drops by $4.2$ on the development set, $7.8$ on AddSent, and $9.1$ on AddOneSent. Finally we find that after only one epoch of training, KAR already achieves an EM of $71.9$ and an F1 score of $80.8$ on the development set, which is even better than the final performance of several strong baselines, such as DCN (EM / F1: $65.4$ / $75.6$) \cite{xiongcaiming:2016} and BiDAF (EM / F1: $67.7$ / $77.3$) \cite{seominjoon:2016}. The above empirical findings imply that general knowledge indeed plays an effective role in KAR. \\
To demonstrate the advantage of our explicit way to utilize general knowledge over the existing implicit way, we compare the performance of KAR with that reported by \citet{weissenborndirk:2017}, which used an encoding-based method to utilize the general knowledge dynamically retrieved from Wikipedia and ConceptNet. Since their best model only achieved an EM of $69.5$ and an F1 score of $79.7$ on the development set, which is much lower than the performance of KAR, we have good reason to believe that our explicit way works better than the existing implicit way.
\begin{table}
\centering
\begin{tabular}
{|m{0.03\linewidth}|m{0.57\linewidth}|m{0.22\linewidth}|}
\hline
\boldmath{$\kappa$} &
\textbf{Average number of inter-word semantic connections per word} &
\textbf{Dev set \newline (EM / F1)} \\
\hline
0 & 0.39 & 74.2 / 82.8 \\
\hline
1 & 0.63 & 74.6 / 83.1 \\
\hline
2 & 1.24 & 75.1 / 83.5 \\
\hline
\textbf{3} & \textbf{2.21} & \textbf{76.7 / 84.9} \\
\hline
4 & 3.68 & 75.9 / 84.3 \\
\hline
5 & 5.58 & 75.3 / 83.8 \\
\hline
\end{tabular}
\caption{\label{t3} With $\kappa$ set to different values in the data enrichment method, we calculate the average number of inter-word semantic connections per word as an estimation of the amount of general knowledge, and evaluate the performance of KAR on the development set.}
\end{table}

\subsection{Model Comparison in the Hunger for Data}
\begin{figure}
\centering
\includegraphics[width=0.98\linewidth]{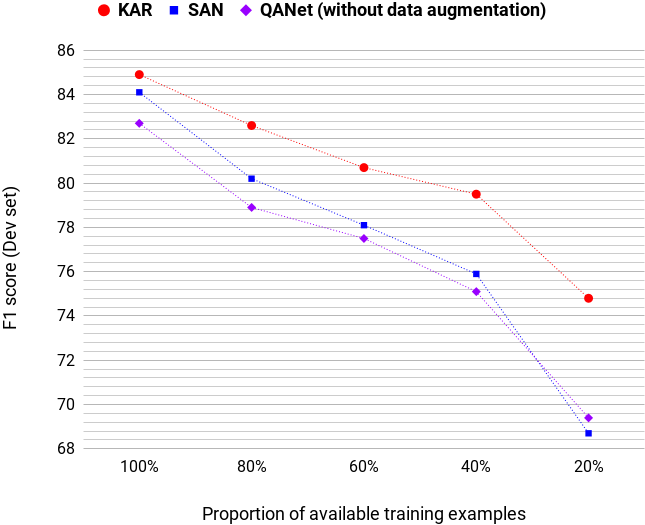}
\caption{\label{f2} With KAR, SAN, and QANet (without data augmentation) trained on the training subsets, we evaluate their performance on the development set.}
\end{figure}
\begin{figure}
\centering
\includegraphics[width=0.98\linewidth]{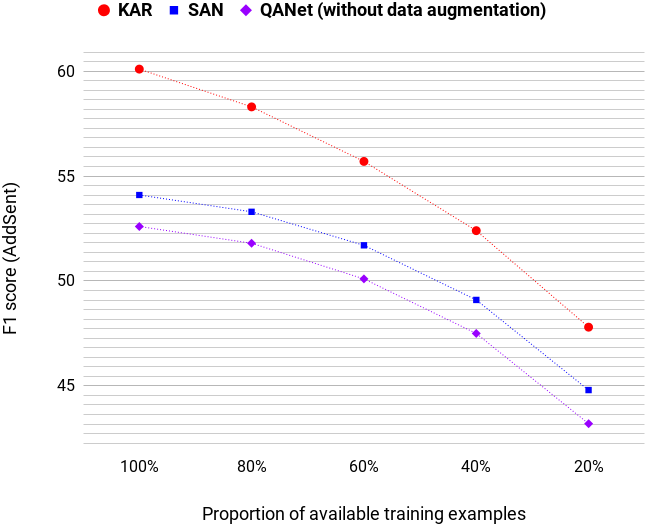}
\caption{\label{f3} With KAR, SAN, and QANet (without data augmentation) trained on the training subsets, we evaluate their performance on AddSent.}
\end{figure}
\begin{figure}
\centering
\includegraphics[width=0.98\linewidth]{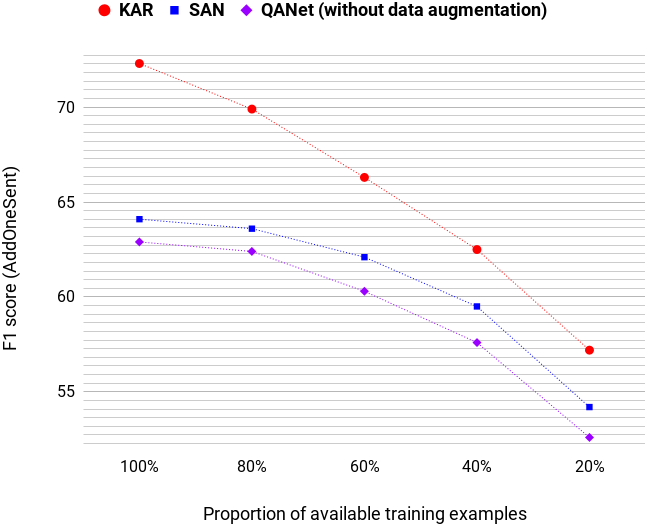}
\caption{\label{f4} With KAR, SAN, and QANet (without data augmentation) trained on the training subsets, we evaluate their performance on AddOneSent.}
\end{figure}

We compare KAR with other MRC models in the hunger for data. Specifically, instead of using all the training examples, we produce several training subsets (i.e. subsets of the training examples) so as to study the relationship between the proportion of the available training examples and the performance. We produce each training subset by sampling a specific number of questions from all the questions relevant to each passage. By separately sampling $1$, $2$, $3$, and $4$ questions on each passage, we obtain four training subsets, which separately contain $20\%$, $40\%$, $60\%$, and $80\%$ of the training examples. As shown in Figure~\ref{f2}, with KAR, SAN (re-implemented), and QANet (re-implemented without data augmentation) trained on these training subsets, we evaluate their performance on the development set, and find that KAR performs much better than SAN and QANet. As shown in Figure~\ref{f3} and Figure~\ref{f4}, with the above KAR, SAN, and QANet trained on the same training subsets, we also evaluate their performance on the adversarial sets, and still find that KAR performs much better than SAN and QANet. That is to say, when only a subset of the training examples are available, KAR outperforms the state-of-the-art MRC models by a large margin, and is still reasonably robust to noise.

\section{Analysis}
According to the experimental results, KAR is not only comparable in performance with the state-of-the-art MRC models, but also superior to them in terms of both the hunger for data and the robustness to noise. The reasons for these achievements, we believe, are as follows:
\begin{itemize}
\item KAR is designed to utilize the pre-extracted inter-word semantic connections from the data enrichment method. Some inter-word semantic connections, especially those obtained through multi-hop semantic relation chains, are very helpful for the prediction of answer spans, but they will be too covert to capture if we simply leverage recurrent neural networks (e.g. BiLSTM) and pre-trained word vectors (e.g. GloVe).
\item An inter-word semantic connection extracted from a passage-question pair usually also appears in many other passage-question pairs, therefore it is very likely that the inter-word semantic connections extracted from a small amount of training examples actually cover a much larger amount of training examples. That is to say, we are actually using much more training examples for model optimization than the available ones.
\item Some inter-word semantic connections are distracting for the prediction of answer spans. For example, the inter-word semantic connection between ``bank'' and ``waterside'' makes no sense given the context ``the bank manager is walking along the waterside''. It is the knowledge aided attention mechanisms that enable KAR to ignore such distracting inter-word semantic connections so that only the important ones are used.
\end{itemize}

\section{Conclusion}
In this paper, we innovatively integrate the neural networks of MRC models with the general knowledge of human beings. Specifically, inter-word semantic connections are first extracted from each given passage-question pair by a WordNet-based data enrichment method, and then provided as general knowledge to an end-to-end MRC model named as Knowledge Aided Reader (KAR), which explicitly uses the general knowledge to assist its attention mechanisms. Experimental results show that KAR is not only comparable in performance with the state-of-the-art MRC models, but also superior to them in terms of both the hunger for data and the robustness to noise. In the future, we plan to use some larger knowledge bases, such as ConceptNet and Freebase, to improve the quality and scope of the general knowledge.

\section*{Acknowledgments}
This work is partially supported  by a research donation from iFLYTEK Co., Ltd., Hefei, China, and a discovery grant from Natural Sciences and Engineering Research Council (NSERC) of Canada.

\bibliography{acl2019}
\bibliographystyle{acl_natbib}

\end{document}